\documentclass[runningheads]{llncs}
\usepackage{booktabs}
\usepackage{hyperref}
\usepackage{colortbl}
\usepackage[table,xcdraw]{xcolor}
\usepackage{amsmath}
\usepackage{amssymb}
\usepackage[T1]{fontenc}
\usepackage{multirow}

\usepackage{graphicx,verbatim}
\usepackage{color}

\usepackage[colorinlistoftodos,textsize=small]{todonotes}
 
\newcommand{\CC}[1]{\cellcolor{gray!50} \textcolor{black}{#1}}
\begin{document}
\title{WASABI: A Metric for Evaluating Morphometric Plausibility of Synthetic Brain MRIs}
\titlerunning{WASABI}

%

\author{
Bahram Jafrasteh\inst{1}$^{*}$\,$^{\dagger}$ \and
Wei Peng\inst{2}$^{*}$ \and
Cheng Wan\inst{3} \and
Yimin Luo\inst{1} \and
Ehsan Adeli\inst{2} \and
Qingyu Zhao\inst{1}
}

\authorrunning{B. Jafrasteh et al.}

\institute{
Department of Radiology, Weill Cornell Medicine, New York, NY, USA \\
\email{\{baj4003, qiz4006\}@med.cornell.edu}
\and
Department of Psychiatry and Behavioral Sciences, Stanford University, Stanford, CA, USA \\
\email{\{wepeng, eadeli\}@stanford.edu}
\and
Department of Electrical and Computer Engineering, Cornell University, Ithaca, NY, USA \\
\email{cw2222@cornell.edu}\\
$^*$Equal contribution. \quad $^\dagger$Corresponding author.
}

\maketitle              

\begin{abstract}
Generative models enhance neuroimaging through data augmentation, quality improvement, and rare condition studies. 
Despite advances in realistic synthetic MRIs, evaluations focus on texture and perception, lacking sensitivity to crucial morphometric fidelity.
This study proposes a new metric, called WASABI (Wasserstein-Based Anatomical Brain Index), to assess the morphometric plausibility of synthetic brain MRIs.
WASABI leverages \textit{SynthSeg}, a deep learning-based brain parcellation tool, to derive volumetric measures of brain regions in each MRI and uses the multivariate Wasserstein distance to compare distributions between real and synthetic anatomies. 
Based on controlled experiments on two real datasets and synthetic MRIs from five generative models, WASABI demonstrates higher sensitivity in quantifying morphometric discrepancies compared to traditional image-level metrics, even when synthetic images achieve near-perfect visual quality. 
Our findings advocate for shifting the evaluation paradigm beyond visual inspection and conventional metrics, emphasizing morphometric fidelity as a crucial benchmark for clinically meaningful brain MRI synthesis.

Our code is available at \url{https://github.com/BahramJafrasteh/wasabi-mri}.


\end{abstract}
\section{Introduction}

Generative models that can synthesize brain MRIs have recently attracted increasing attention in neuroimaging studies as they have the potential to aid in disease progression prediction, counterfactural generation, clinical education, and data augmentation~\cite{sun2022hierarchical,ali2022role,chintapalli2024generative,tudosiu2024realistic}. Despite substantial efforts devoted to designing generative architectures, how to effectively evaluate the quality of generated samples remains underexplored in the neuroimaging domain.

Early brain MRI generative models, primarily based on VAEs or GANs \cite{kwon2019generation,ali2022role}, often produced low-resolution, blurry, noisy, and artifact-prone images due to algorithmic and computational limitations. To quantitatively assess image quality, studies often employed metrics commonly used in the computer vision community, such as Fréchet Inception Distance (FID) \cite{heusel2017gans}, Multi-Scale Structural Similarity Index (MS-SSIM) \cite{wang2003multiscale}, and Maximum Mean Discrepancy (MMD) \cite{gretton2012kernel}. 
These metrics quantify differences between real and synthetic images based on distributional, textural, and perceptual characteristics, often in a feature space. 
These metrics are often adopted to support conclusions drawn from visual inspection. In fact, given the general poor quality of early synthetic MRIs, visual inspection alone can easily distinguish real from fake MRIs and gauge realism between models \cite{yu2019ea}. \\
With recent advances in generative models, studies have gradually converged to producing brain MRIs with ``near-perfect'' visual quality \cite{peng2024metadata,xu2024medsyn,pinaya2022brain}. Subtle anatomical inaccuracies from these synthetic MRIs become harder to detect by human experts or existing metrics. For example, in a user study \cite{peng2024metadata}, neuroradiologists were only able to distinguish between real and fake MRIs 70\% of the time, and they were no longer able to reliably judge which generative models produce more realistic brain MRIs. This progress calls for a paradigm shift in evaluation protocols. Beyond measuring image quality, model evaluation should assess how the generated images reflect true anatomy (e.g., whether the cortical thickness of the frontal region falls in a typical distribution of real data). 
We argue that visual inspection and traditional metrics lack sensitivity to morphometric differences. Instead, a new summary statistic that focuses on morphometric fidelity should be adopted to guide the development of generative models toward producing clinically meaningful synthetic MRIs.\\
To achieve this goal, we propose an efficient metric called Wasserstein-Based Anatomical Brain Index (WASABI). 
To measure morphometric discrepancy between two sets of MRIs, WASABI first applies SynthSeg, a deep learning brain-parcellation tool,  to derive regional volumetric measures in each MRI. Then, WASABI measures the multivariate Wasserstein distance between the two distributions of high-dimensional brain measures. To test its sensitivity in detecting morphometric distances between MRI datasets in a controlled setting, we create 4 ways of partitioning real MRIs of the Alzheimer’s Disease Neuroimaging Initiative (ADNI) dataset, where we know the ground-truth ranking of morphometric discrepancies across partition scenarios. Compared to traditional metrics, only WASABI aligns with the correct rank with well stratified morphometric distances across data partition scenarios. Finally, based on evaluating distances across two real MRI datasets and five generative models, only WASABI indicates the real MRIs have higher morphometric fidelity compared to synthetic MRIs, whereas other distance metrics tend to be biased by discrepancy in image appearance.

\section{Methods}
\subsection{Common Metrics for Evaluating Synthetic MRIs}

Synthetic MRI quality metrics measure a certain type of distance between synthetic and real data.
For example,
\textbf{FID} (Fréchet Inception Distance) \cite{heusel2017gans} compares the distribution of features between real and synthetic images. It uses a pretrained Inception network to extract features, models them as multivariate Gaussians, and computes the Fréchet distance between these two distributions, with lower scores indicating higher similarity and better image quality.
\textbf{MS-SSIM} (Multi-Scale Structural Similarity Index) \cite{wang2003multiscale}, on the other hand, evaluates the structural similarity between images at multiple scales by analyzing luminance, contrast, and texture. Higher values indicate greater perceptual similarity, making it useful for assessing fine details in medical images. Lastly, 
\textbf{MMD} (Maximum Mean Discrepancy) \cite{gretton2012kernel} is a statistical measure that compares the distributions of two datasets by computing the difference in their mean embeddings in a reproducing kernel Hilbert space. Lower values suggest that the generated images better align with the real data distribution. All the above metrics measure image-level or representation-level distances but do not assess anatomical fidelity.\\
\begin{figure}[!t]
\centering

\includegraphics[width=1\textwidth,keepaspectratio]{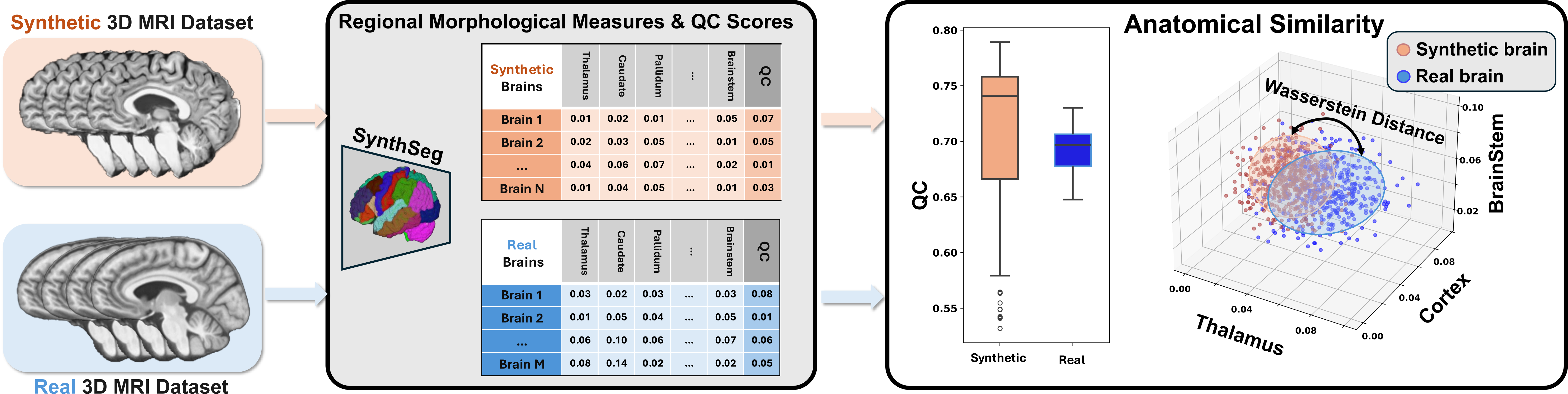}
\caption[]{Wasserstein-Based Anatomical
Brain Index (WASABI): To measure the anatomical fidelity of a set synthetic MRIs, we compute their distance to a reference dataset of real MRIs by applying SynthSeg to derive regional morphological measures of the two datasets. Then WASABI is computed as the Wasserstein distance between the two multivariate distributions of brain measures. The quality control scores derived by SynthSeg are used to identify low quality images.}\label{fig:fig1_overview}
\end{figure}
\subsection{Quality Metric Based on Morphometric Similarity}
Now we introduce a simple and efficient metric, WASABI, the first metric explicitly designed to assess whole-brain morphometric plausibility of synthetic brain MRIs. WASABI measures ``anatomical distance'' by comparing a set of morphometric features between a set of generated images and a reference real dataset (Figure \ref{fig:fig1_overview}). In particular, we use SynthSeg \cite{billot2023synthseg}, a pre-trained deep learning tool, to derive volumetric measures associated with 68 cortical regions and 32 subcortical regions in each MRI. We choose SynthSeg over traditional MRI processing pipelines because it is optimized for efficient processing (less than 30 seconds processing time using GPU compared to 10+ hours of processing time by traditional Freesurfer). Another key advantage of SynthSeg lies in its contrast-invariant training, which enhances its adaptability across different MRI sequences and scanner types. 
For simplicity, we average the left and right hemisphere measurements, resulting in 52 bilateral volumetric measures
(although our approach can be easily extended to other types of measures like curvature and cortical thickness). To compare these regional measures between real and synthetic data, prior works have used univariate Cohen's $d$ to measure the discrepancy in each brain region separately \cite{peng2024metadata,wu2024evaluating}. This approach results in many separate distance scores and cannot provide a unified whole-brain summary or capture multivariate anatomical coherence.\\
Here, we propose to generate a holistic scalar summary score to intuitively quantify the morphometric distance in a high-dimensional space. Specifically, we first normalize each regional volume measure by the total intracranial volume to remove the effect of head size.
Let $\mathbf{X}\in \mathbb{R}^{N \times 52}$ be the set of normalized morphometric measures derived from the $N$ synthetic MRIs and $\mathbf{Y}\in\mathbb{R}^{M \times 52}$ be the set of normalized morphometric measures obtained from a reference dataset of $M$ real MRIs, we assume that both $\mathbf{X}$ and $\mathbf{Y}$ follow Gaussian distributions, i.e., \sloppy $\mathbf{X} \sim \mathcal{N}(\mu_X, \Sigma_X)$ and $\mathbf{Y} \sim \mathcal{N}(\mu_Y, \Sigma_Y)$, where $\mu_X$ and $\mu_Y$ represent the mean of the two distributions, and $\Sigma_X$ and $\Sigma_Y$ denote their covariance matrices. To quantify the morphometric similarity between these two distributions, we compute the squared Wasserstein distance \cite{villani2009optimal} between their respective multivariate Gaussian approximations:
\begin{equation}
    W_2^2(\mathbf{X}, \mathbf{Y}) = \|\mu_X - \mu_Y\|_2^2 + \text{Tr} \left(\Sigma_X + \Sigma_Y - 2 \left(\Sigma_X^{1/2} \Sigma_Y \Sigma_X^{1/2} \right)^{1/2} \right),
\end{equation}
where $\text{Tr}(\cdot)$ refers to the matrix trace operator. Compared with other statistical distances (e.g., KL or JS divergence), Wasserstein distance is a true metric, i.e., symmetric and satisfying triangle inequality. 

By assuming Gaussian distributions as in FID~\cite{heusel2017gans}, the computation of the Wasserstein distance becomes extremely efficient, as it reduces the problem to simple matrix operations without requiring iterative numerical optimization of the optimal transport problem (\(\mathcal{O}(d^3)\) complexity for computation of the square root of the covariance matrix with dimension $d$). This makes the distance calculation both fast and computationally scalable, especially for large-scale datasets. Lastly, SynthSeg also produces a QC score indicating the reliability and accuracy of the gray matter segmentation. 
Based on our experimental results, we explain in the next section how to leverage the QC score and our proposed WASABI metric to generate insights into the quality of synthetic MRIs.







\section{Experimental Configurations} \label{Exp:conf}
This section describes our experiments on testing WASABI using two real MRI datasets and five state-of-the-art (SOTA) brain MRI generative models. 

\subsection{Real Data Experiments on ADNI.}
\label{sec:config1}
Evaluating the validity of an MRI-quality metric is inherently challenging as there is no ground-truth in knowing which generative model is morphometricly more similar to real data. Therefore, before investigating the morphometric plausibility of synthetic MRIs, we first validated the soundness of our metric in a controlled scenario based on real data. We utilized T1-weighted MRI data from ADNI 1, 2, 3, and GO cohorts ~\cite{petersen2010alzheimer}.
All scans underwent skull stripping and segmentation using SynthSeg \cite{billot2023synthseg}, and transformed into the MNI standard coordinate system \cite{brett_etal_2002_mni}. A total of 11,436 T1 images from 1,395 patients (696 males and 699 females, aged 55 to 91.4 years) successfully passed through our pipeline. For this analysis, we removed 180 images with a gray-matter QC score below 0.7 (below the 1.5x interquartile range from the first quartile) to ensure the credibility of SynthSeg measures. Note, we removed these low quality images because they were trivial to detect both visually and by all metrics. We focused the following analysis on high-quality synthetic images where human assessment becomes unreliable and a good quantitative metric is most needed. 

Next, we constructed two subsets of images from ADNI 
in 4 different scenarios, measured the distance between the two subsets in each scenario, and ranked the distance across the 4 scenarios.
The first scenario was to separate the dataset based on sex (Males vs. Females). The second was to compare images from normal controls with images with mild cognitive impairment (NC vs. MCI). The third was to compare NC subjects to individuals with Alzheimer's disease (NC vs. AD). Lastly, as a reference comparison, we randomly split the NC images into two subsets and measured the within-cohort distance (NC vs. NC). These controlled subsets allow us to examine how well a metric captures meaningful morphometric variations before applying it to synthetic data. Specifically, given the nature of the aging and cognitive impairment, we anticipate that the distance between NC vs. NC would be the smallest and the distance between NC vs. AD would be the largest. Based on the effect sizes reported in the prior literature \cite{guo2010voxel,ruigrok2014meta}, we also hypothesize that the distance between males and females falls between NC vs. MCI and NC vs. AD. 

Given any two subsets of MRIs from ADNI, we randomly sampled 500 images from either subset (so that our distance estimate was not biased by sample size variations) and computed WASABI, MS-SSIM, FID, and MMD. Similar to~\cite{tudosiu2024realistic},
we used the 101-layer version of MedicalNet3D \cite{chen2019med3d} as the feature extractor for FID and MMD computation in all experiments. In each of the four comparison scenarios, we repeated the above random sampling 1000 times to generate a distribution for the distances. 

\subsection{Comparing Morphometric Realism of Generative Models}
Next, we assessed the effectiveness of the proposed metric in evaluating five SOTA brain generative models. After a careful literature review and extensive testing of existing models, we identified 5 models that can generate high-resolution 3D MRI volumes with good visual quality (see Figure.~\ref{fig:fig2_views}). We excluded models that only generate low-resolution 
or with poor image quality (blurry, noisy, or with artifacts). Those synthetic MRIs are obviously anatomically inaccurate and cannot be successfully used by any processing pipeline (including SynthSeg). To make the comparison fair, we only focused on image generation from scratch and excluded conditional generative models that generate samples based on an existing real MRI \cite{dar2019image,puglisi2024enhancing}. The chosen models are:\\
\underline{\textbf{Latent Diffusion Models (LDM)}}: LDM was trained on the UKbiobank (UKB) dataset ~\cite{biobank2016,biobank2017}. We utilized 1000 synthetic MRIs with age $>50$ years randomly sampled from the 100,000 Synthetic T1 images released by \cite{pinaya2022brain}. \\
\underline{\textbf{Hierarchical Amortized GAN (HA-GAN)}} \cite{sun2022hierarchical}: HA-GAN employs a dual-branch generator that combines low-resolution full-volume synthesis with stochastic high-resolution sub-volume sampling. We used a released model pre-trained on the GPS dataset \cite{holmes2015brain} to generate 1000 synthetic MRIs.\\
\underline{\textbf{BrainSynth} \cite{peng2024metadata}}:
BrainSynth is a metadata-conditioned generative model designed to synthesize anatomically plausible 3D brain MRIs by incorporating subject-specific attributes, ensuring realistic structural variability. It was trained on a multi-site dataset, including the ADNI \cite{petersen2010alzheimer} and the National Consortium on Alcohol and Neurodevelopment in Adolescence (NCANDA) \cite{brown2015national}. We used their released pre-trained model to generate 1000 MRIs.\\
\underline{\textbf{MedSyn} \cite{xu2024medsyn}}: MedSyn is a generative model that synthesizes high-quality 3D CT images based on textual descriptions while incorporating anatomical awareness to preserve structural accuracy. MedSyn was originally trained on CT images. We used a variant of the model for MRI synthesis pre-trained on the same multi-site dataset as BrainSynth.\\
\underline{\textbf{Med-DDPM} \cite{dorjsembe2024conditional}}: Med-DDPM is a diffusion-based generative model designed for high-resolution 3D medical image synthesis. It utilizes a cascaded denoising process to iteratively refine synthetic MRI volumes, ensuring high anatomical fidelity and structural consistency. Med-DDPM was trained using unnormalized clinical brain MRI without skull stripping. It requires whole-head masks to generate a synthetic MRI. We used a pre-trained version of Med-DDPM to generate 1000 synthetic MRI volumes from randomly selected masks.\\
Given that the above models were trained on different datasets, we incorporate another real dataset so that we can evaluate the distance between real and synthetic data and between the two real datasets. Specifically, we randomly selected 1000 MRIs from UKB~\cite{biobank2016,biobank2017}, which consists of high-quality T1-weighted MRI scans \cite{alfaro2018image} collected from a diverse cohort of participants across the UK aged between 40 and 69 years-old. Finally, we used the same data processing procedures to derive the distribution of FID, MMD, MS-SSIM, and WASABI between synthetic and UKB data (i.e., 500 real vs. 500 synthetic MRIs, repeated 1000 times). Lastly, we also computed the distance between the two real datasets (UKB vs. normal controls in ADNI) as well as the within-UKB distance (distance between two halves of the UKB samples split randomly). 

\section{Results and Discussion}
\textbf{Results on ADNI}. 
\begin{figure}[!t]
\centering

\includegraphics[width=1.\textwidth,keepaspectratio]{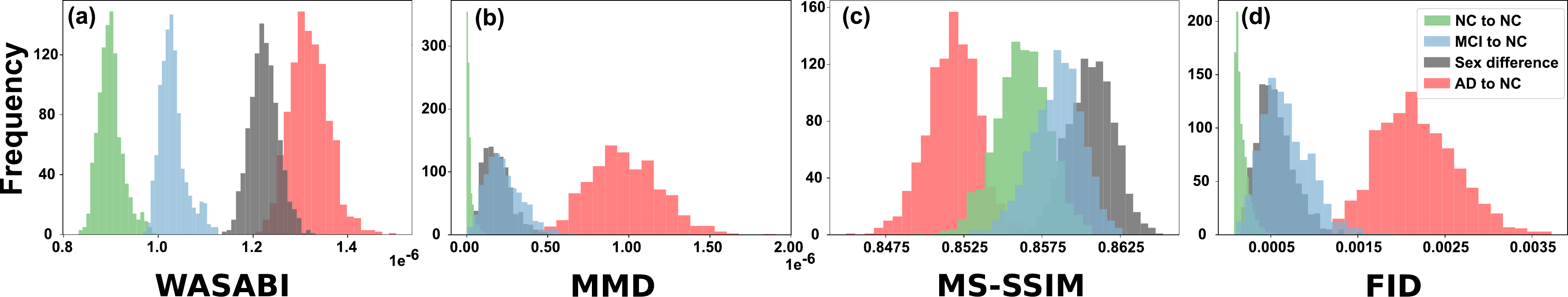}
\caption[]{Distribution of  WASABI (ours), MMD, MS-SSIM, and FID between two subsets of real MRIs of ADNI separated by AD vs. NC, males vs. females (sex difference), MCI vs. NC, and NC vs. NC (two random subsets of NC).}\label{fig:fig3_dist_adni}
\end{figure}
Figure \ref{fig:fig3_dist_adni}a-d shows the distribution of FID, MMD, MS-SSIM, and WASABI for measuring the morphometric similarity in the four comparison scenarios described in Section~\ref{sec:config1}.
An effective metric should be capable of clearly distinguishing between distributions, with large separation reflecting the increasing morphometric differences across the groups. Aligning with our expectation, only our metric, WASABI, shows a clear separation between the distributions. The within-NC distances, regarded as the reference ``null distribution''  of the metric, were the smallest as the difference was only due to the random split of a homogeneous cohort. The AD-NC distance shows the greatest divergence from this reference value, suggesting a severe anatomical deterioration associated with brain atrophy. Meanwhile, the morphometric distances associated with MCI and sex difference were moderate but still detectable (significantly larger than the reference distribution of NC-NC).

The other three metrics were not as sensitive as ours in identifying morphometric differences. For FID and MMD, the distributions of MCI-NC and sex difference significantly overlap with the reference distribution of NC-NC
, suggesting that overall the MCI and sex effects on morphometric changes were not detectable based on these two metrics. 
Lastly, MS-SSIM introduced more overlap between the four distributions than WASABI and incorrectly ranked the distributions: the MCI-NC distance and between-sex distance were smaller (higher similarity) than the reference NC-NC distance, violating anatomical reality. These results collectively show that WASABI was the only metric that reliably measured morphometric differences between cohorts and, therefore, useful for examining the morphometric fidelity of synthetic MRIs.

\textbf{Results on Generative Models}. 
\begin{figure}[!t]
\centering

\includegraphics[width=1\textwidth,keepaspectratio]{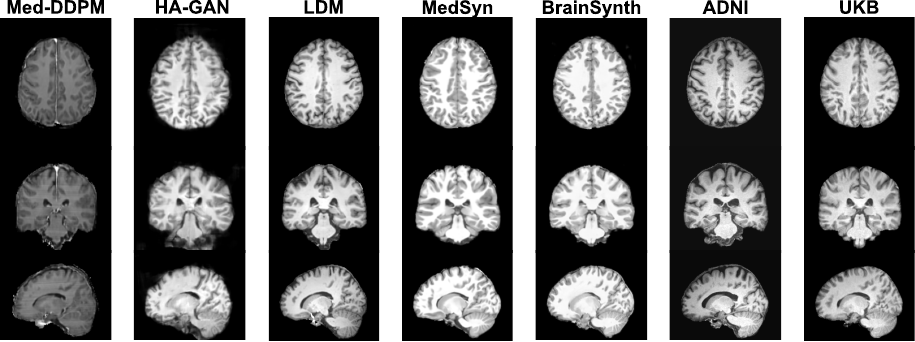}
\caption[]{Axial, Coronal, and Sagittal views of a random sample from five generative models and two real datasets (ADNI and UKB).}\label{fig:fig2_views}
\end{figure}
Figure \ref{fig:fig2_views} visualizes samples generated by the five models, confirming that the synthetic MRIs are visually realistic compared to real data in ADNI and UKB. Examining the QC scores of all images (Table \ref{tab:metrics_comparison}) revealed that BrainSynth and MedSyn could synthesize MRIs with equal or even better quality than the real data of ADNI, indicating the need for further anatomical assessment beyond perceptual assessment. In practice, we found that when the QC score was low, it became trivial for all metrics to identify difference between real and synthetic MRIs. We, therefore, only used images with QC $>$ 0.7 (the threshold used for the ADNI analysis) from each dataset to derive the distance metrics. Med-DDPM was omitted in the following analysis as most samples failed QC.
According to Table \ref{tab:metrics_comparison}, our proposed WASABI was the only metric suggesting the real data in ADNI had the smallest distance to UKB while all synthetic MRIs were less realistic in terms of morphometric similarity. The worse metric recorded for LDM and HA-GAN also aligned with prior findings in \cite{peng2024metadata,wu2024evaluating}. Not surprisingly, generative models with higher QC scores tended to have better WASABI scores. On the other hand, FID and MMD yielded similar results, showing that the distance between synthetic samples from LDM and the real UKB MRIs was even smaller than the distance between two real datasets. Given that LDM was the only model trained on UKB, our results suggest that instead of focusing on morphometric realism, FID and MMD might only capture differences in image perceptual representations, which were largely biased by site/scanner differences. Lastly, MS-SSIM could not effectively stratify the quality of different generative models and incorrectly indicated that the real ADNI data had the largest distance to UKB compared to synthetic MRIs.    

\begin{table}[htbp!]
\centering
\renewcommand{\arraystretch}{1.2}
\setlength{\tabcolsep}{5pt}
\caption{Left: SynthSeg QC scores of real or synthetic MRIs; Right: MS-SSIM, FID, MMD, and WASABI between each dataset and UKB. The first row (highlighted in gray) records within-UKB distance, regarded as the reference `null' value.}
\resizebox{\textwidth}{!}{%
\begin{tabular}{l|c|cccccccc}

\toprule
\textbf{Dataset} & \textbf{QC} & \multicolumn{4}{c}{\textbf{Distance w.r.t. UKB} } \\


 & & \multicolumn{1}{c}{\textbf{MS-SSIM} } & \multicolumn{1}{c}{\textbf{FID} } & \multicolumn{1}{c}{\textbf{MMD} } & \multicolumn{1}{c}{\textbf{WASABI (Ours)} } \\
\midrule
UKB        & \textbf{0.78(0.01)}
 & \CC{0.962(0.000)}      & \CC{0.04(0.029)}  & \CC{0(0.003)}  & \CC{0.1(0.019)}  \\
ADNI (NC)        & 0.73(0.02) & 0.879(0.001)  & 19.88(0.898)  & 8.03(0.301)  & \textbf{2.39(0.176)}  \\
\midrule
MedSyn    & 0.77(0.02) & \textbf{0.889(0.001)}  & 27.33(0.569)  & 13.15(0.282)  & 3.29(0.274)  \\

BrainSynth& 0.77(0.01) & 0.885(0.001)  & 30.98(0.418)  & 14.78(0.193)  & 5.03(0.33)   \\
LDM       & 0.711(0.02) & 0.888(0.001)  & \textbf{11.08(0.444)}  & \textbf{5.16(0.197)}  & 12.26(0.504)  \\
HA-GAN     & 0.66(0.03) & 0.882(0.000)      & 60.13(0.412)  & 29.23(0.199)  & 22.63(0.461)  \\
Med-DDPM     & 0.61(0.08) & --      & --  & --  & -- \\
\bottomrule
\end{tabular}
}
\label{tab:metrics_comparison}
\end{table}

\section{Conclusion}

In this study, we proposed a novel metric for evaluating the anatomical realism of synthetic brain MRIs generated by advanced generative models. While traditional image-level metrics like FID, MS-SSIM, and MMD have been commonly used, they fall short in assessing the true morphometric fidelity needed for clinical applications. As visual quality improves, it is essential to focus on how well synthetic MRIs reflect real brain anatomy. We introduced WASABI, a more objective and scalable metric for quantifying morphometric realism and guiding generative model development. Our results suggest that WASABI can better capture subtle morphometric differences and help advance the creation of synthetic MRIs with real clinical utility. Several limitations should be noted. First, the WASABI metric assumes Gaussianity of brain volumetric measures. Although this assumption was confirmed on the UKB data by the Henze–Zirkler test \cite{henze1990class} (p-value = 0.24), its generalizability needs to be tested on other datasets and brain measures. Second, to increase clinical relevance, WASABI needs to be tested and extended to MRIs with pathology, e.g., patients with severe brain atrophy or tumors.
\section*{Acknowledgments.}
This work was partially supported by NIH R00 AA028840, R01 AG089169, BBRF Young Investigator Grant, and Stanford University HAI Hoffman-Yee Grant.

\section*{Disclosure of Interests}
The authors declare that they have no competing interests.
\bibliographystyle{splncs04}
\bibliography{ref}

\end{document}